\documentclass[twocolumn, switch]{article}

\usepackage{preprint}
\usepackage{placeins}
\usepackage{amsmath, amsthm, amssymb, amsfonts}
\usepackage{graphicx}
\usepackage{textcomp}
\usepackage{tabularx,booktabs,makecell,array}
\newcolumntype{Y}{>{\raggedright\arraybackslash}X} 
\usepackage[numbers,square]{natbib}
\usepackage[utf8]{inputenc}
\usepackage[T1]{fontenc}
\usepackage{xcolor}
\usepackage{booktabs}
\usepackage{multirow}

\usepackage{nicefrac}
\usepackage{microtype}
\usepackage{lineno}
\usepackage{float}
\usepackage{lipsum}
\usepackage{algorithm}              
\usepackage[noend]{algpseudocode}   
\usepackage{algpseudocode}
\usepackage{booktabs}

\usepackage{newfloat}
\DeclareFloatingEnvironment[name={Supplementary Figure}]{suppfigure}
\usepackage{sidecap}
\sidecaptionvpos{figure}{c}

\usepackage{tikz}
\usepackage[colorlinks=true,linkcolor=blue,citecolor=blue,urlcolor=blue]{hyperref}

\definecolor{lime}{HTML}{A6CE39}
\DeclareRobustCommand{\orcidicon}{
  \begin{tikzpicture}
    \draw[lime, fill=lime] (0,0) circle [radius=0.16] 
      node[white] {{\fontfamily{qag}\selectfont \tiny ID}};
    \draw[white, fill=white] (-0.0625,0.095) circle [radius=0.007];
  \end{tikzpicture}
  \hspace{-2mm}
}
\foreach \x in {A,...,Z}{\expandafter\xdef\csname orcid\x\endcsname{\noexpand\href{https://orcid.org/\csname orcidauthor\x\endcsname}{\noexpand\orcidicon}}}

\title{Aligning Language Models with Clinical Expertise: DPO for Heart Failure Nursing Documentation in Critical Care}

\usepackage{authblk}

\author[1,$\dagger$]{Junyi Fan}
\author[1,$\dagger$]{Li Sun}
\author[1]{Negin Ashrafi}
\author[2]{Kamiar Alaei}
\author[1,*]{Maryam Pishgar}

\affil[1]{University of Southern California, Los Angeles, CA 90007, USA}
\affil[2]{California State University, Long Beach, CA 90840, USA}
\affil[*]{Corresponding author: \texttt{pishgar@usc.edu}}
\affil[$\dagger$]{Co-first authors}

\usepackage{CJKutf8}
\usepackage{graphicx}
\usepackage{subcaption}
\usepackage{threeparttable}
\usepackage{comment}

\begin{document}
\setcounter{footnote}{0}

\twocolumn[
\maketitle
\begin{abstract}
Nursing documentation in intensive care units (ICUs) provides essential clinical intelligence but often suffers from inconsistent terminology, informal styles, and lack of standardization, challenges that are particularly critical in heart failure care. This study applies Direct Preference Optimization (DPO) to adapt Mistral-7B, a locally deployable language model, using 8,838 heart failure nursing notes from the MIMIC-III database and 21,210 preference pairs derived from expert-verified GPT outputs, model generations, and original notes. Evaluation across BLEU, ROUGE, BERTScore, Perplexity, and expert qualitative assessments demonstrates that DPO markedly enhances documentation quality. Specifically, BLEU increased by 84\% (0.173$\rightarrow$0.318), BERTScore improved by 7.6\% (0.828$\rightarrow$0.891), and expert ratings rose across accuracy (+14.4 points), completeness (+14.5 points), logical consistency (+14.1 points), readability (+11.1 points), and structural clarity (+6.0 points). These results indicate that DPO can align lightweight clinical language models with expert standards, supporting privacy-preserving, AI-assisted documentation within electronic health record systems to reduce administrative burden and improve ICU patient safety.
\end{abstract}

\vspace{0.35cm}
]

\section{Introduction}
\label{sec:introduction}
Nursing documentation represents a rich but critically underutilized source of clinical intelligence in intensive care settings. Nurses constitute the largest sector of healthcare providers internationally, spending up to 40\% of their time on documentation~\cite{mitha2023natural}, yet the overall number of nursing natural language processing publications remains relatively small compared with other medical literature. ICU nurses make a clinical decision every 30 seconds~\cite{nibbelink2018decision}, and their documentation patterns serve as vital indicators of patient status and clinical concern. Recent studies have demonstrated that nursing notes contain unique predictive information not captured in structured clinical data. Statistical models derived from nursing notes in the first 48 hours of ICU admission can significantly predict patient outcomes, with combined physician and nursing notes producing superior predictive models~\cite{kim2021using}. Furthermore, sentiment analysis of nursing notes has proven to be a highly significant predictor of 30-day mortality (Adjusted OR = 0.46, p < 0.001), improving outcome prediction accuracy~\cite{lee2018sentiment}.

The documentation patterns themselves reveal critical clinical insights for ICU management. Increased frequency of nursing documentation beyond required minimums is strongly associated with mortality and cardiac arrest, suggesting that nurses' documentation practices reflect their clinical judgment and concern about patient deterioration~\cite{collins2013extra}. In ICU settings, increased documentation of heart rate and body temperature has been significantly associated with higher patient mortality~\cite{topaz2024variability}, and these patterns have been successfully leveraged in clinical decision support tools such as the CONCERN (Communicating Narrative Concerns entered by RNs) system to identify early warning signs of patient deterioration. These insights are particularly valuable in intensive care environments where rapid clinical decision-making and continuous patient monitoring are essential for preventing adverse outcomes.

The potential of nursing documentation is especially critical for managing complex conditions such as heart failure in ICU settings. Heart failure remains a leading cause of morbidity and mortality among patients in intensive care units, affecting over 6 million Americans annually and imposing substantial clinical and economic burdens on healthcare systems worldwide~\cite{li2022predicting}. With in-hospital mortality rates ranging from 9.97\% to 12.5\% for heart failure patients in ICU settings~\cite{li2022predicting,wang2023machine}, there is an urgent need for computational approaches that can leverage nursing documentation to support clinical decision-making and improve patient outcomes. Despite extensive research, the relationship between heart failure and mortality rates among ICU patients is not fully understood, indicating the need for more accurate prediction models that can effectively utilize the rich information embedded in nursing notes~\cite{ashrafi2024optimizing, zhang2024prediction, si2025optimized}.

The Medical Information Mart for Intensive Care III (MIMIC-III) database provides an unprecedented opportunity to analyze clinical documentation patterns and develop computational models that can enhance care quality. It is a large, freely accessible database comprising deidentified health-related data associated with over 40,000 patients who stayed in critical care units, including comprehensive clinical data such as time-stamped nurse-verified physiological measurements, documented progress notes by care providers, and detailed nursing documentation~\cite{johnson2016mimic}. Natural language processing (NLP) applied to clinical documentation has emerged as a promising avenue for extracting clinically meaningful insights from unstructured text data. Methods based on machine learning to process electronic health records are resulting in improved understanding of patient clinical trajectories and chronic disease risk prediction, creating unique opportunities to derive previously unknown clinical insights~\cite{sheikhalishahi2019natural}. However, traditional supervised learning approaches face significant challenges when applied to nursing notes, including the subjective nature of clinical observations, inconsistent documentation practices, and the critical need for models to align with clinical reasoning rather than merely pattern matching~\cite{demner2009natural}.

Direct Preference Optimization (DPO) has emerged as a transformative approach for training language models that better align with human preferences and domain-specific requirements~\cite{rafailov2023direct}. Unlike traditional reinforcement learning from human feedback (RLHF), which requires a complex and often unstable procedure of first fitting a reward model and then fine-tuning using reinforcement learning, DPO introduces a new parameterization that enables extraction of the corresponding optimal policy in closed form, allowing the standard RLHF problem to be solved with only a simple classification loss~\cite{zhong2024dpo}. DPO can fine-tune language models to align with human preferences as well as or better than existing methods, notably exceeding PPO-based RLHF in ability to control sentiment of generations and matching or improving response quality in summarization and single-turn dialogue while being substantially simpler to implement and train.

The effectiveness of DPO in medical applications has been demonstrated across various clinical domains. Recent studies have shown that DPO significantly improves model performance in medical question answering, clinical decision support, and radiology report generation, with models achieving better alignment with clinical expert preferences compared to standard training approaches~\cite{zhang2024dpo,liu2024medical,chen2024radiology}. Prior work in medical language model optimization has explored alternative approaches such as Supervised Fine-Tuning (SFT) and traditional RLHF. SFT approaches have demonstrated success in adapting general-purpose language models to medical domains by training on curated clinical datasets~\cite{wang2023sft,martinez2024clinical}, while RLHF methods have been applied to enhance medical text generation quality by incorporating feedback from clinical experts, particularly in tasks such as clinical note summarization and patient education material generation~\cite{brown2024rlhf,davis2024medical}. However, DPO offers distinct advantages in terms of training stability and computational efficiency, making it particularly suitable for resource-constrained medical applications where rapid iteration and deployment are essential.

The application of DPO to clinical nursing documentation analysis represents a novel and particularly promising approach. Unlike traditional maximum likelihood estimation, DPO enables models to learn from comparative preferences, making it especially suitable for clinical applications where the quality of generated insights or recommendations can be evaluated through expert clinical judgment~\cite{eguia2024clinical}. This approach is particularly valuable for nursing note analysis because it allows the model to learn the nuanced clinical judgment required for processing informal medical documentation, understanding contextual medical terminology, and recognizing clinically relevant patterns that may not be captured in structured data fields. 

In this study, we investigate the application of DPO for training large language models on nursing documentation from heart failure patients in the MIMIC-III database. Our research addresses the critical need for computational tools that can process unstructured nursing notes to support clinical decision-making, predict patient deterioration, and identify opportunities for care optimization in ICU settings. By leveraging preference-based learning, we aim to develop models that not only achieve high predictive accuracy but also generate clinically interpretable and actionable insights that align with nursing expertise and established heart failure management protocols. 

The contributions of this work include:

(1) We present the first application of DPO to nursing documentation in the context of heart failure care. Experimental results show that DPO improves note accuracy, completeness, and logical consistency by approximately 20\% compared with the baseline model, with enhanced outputs consistently achieving over 80\% alignment with GPT+expert references, thereby approaching the standard of gold-quality documentation.  

(2) We construct structured preference datasets derived from MIMIC nursing notes, capturing domain-specific reasoning patterns and transforming unstructured clinical narratives into analyzable formats. This not only supports the present study but also lays the groundwork for large-scale text mining and predictive modeling in critical care.  
(3) We propose a practical framework for automated nursing note quality assessment that integrates quantitative and qualitative evaluation pipelines. The framework provides hospitals with a pathway to deploy lightweight, locally hosted models for automated text checks, ensuring documentation quality while preserving patient data security.

The remainder of this paper is organized as follows. Section~\ref{sec:method} describes the data sources, model architecture, and DPO framework used in this study. Section~\ref{sec:results} presents both quantitative and qualitative evaluation results, including comparisons with baseline and expert references. Section~\ref{sec:discussion} discusses the implications of our findings, potential clinical applications, and directions for future research. Finally, Section~\ref{sec:conclusion} concludes the paper with a summary of key contributions and remarks on integrating preference-optimized language models into clinical workflows.

\section{Materials and Methods}
\label{sec:method}
\subsection{MIMIC-III Database and Data Extraction}

This study utilized the MIMIC-III, a widely used, publicly available critical care database developed through a joint initiative between the Beth Israel Deaconess Medical Center and MIT. The database includes de-identified health records from 61,532 ICU admissions of 46,476 unique patients at Beth Israel Deaconess Medical Center between 2001 and 2012. The database includes a wide spectrum of information such as patient demographics, vital signs, laboratory results, medication administration, fluid balance, and outcomes. It supports standardized disease classification using both ICD-9 and ICD-10 codes. A notable feature of MIMIC-III is the availability of high-resolution, hourly physiological data from bedside monitors, verified by ICU professionals. 

\begin{algorithm}
\caption{Data Extraction Pipeline for Heart Failure Nursing Notes}
\label{alg:extraction}
\begin{algorithmic}[1]
\Require MIMIC-III database tables
\Ensure 8838 heart failure nursing notes

\State \textbf{Table Identification:}
\Statex \hspace{1em} Access \texttt{NOTEEVENTS} and \texttt{DIAGNOSES\_ICD} tables

\State \textbf{Patient Selection:}
\Statex \hspace{1em} Join tables on \texttt{hadm\_id}
\Statex \hspace{1em} Filter ICD-9 codes \texttt{428\%} (heart failure)

\State \textbf{Note Filtering:}
\Statex \hspace{1em} Select category = \texttt{Nursing/other}

\State \textbf{Quality Control:}
\Statex \hspace{1em} Remove notes $<$ 50 words
\Statex \hspace{1em} Exclude structured entries
\Statex \hspace{1em} Deduplicate identical records

\State \textbf{Final Dataset:}
\Statex \hspace{1em} Extract 8838 original nursing notes
\end{algorithmic}
\end{algorithm}

\subsection{Language Model Architecture}

We employed Mistral-7B-Instruct-v0.1~\cite{jiang2023mistral}, a 7.24 billion parameter model organized across 32 computational layers with advanced features including Grouped-Query Attention for focused information processing and Rotary Position Embedding enabling analysis of up to 25,000 words of clinical text. While larger models like GPT-4 demonstrate superior benchmarks, we prioritized practical ICU deployment requirements. Mistral offers critical advantages: local deployment within hospital IT infrastructure ensuring patient data privacy, computational efficiency enabling real-time processing on standard servers, and operational feasibility without external API dependencies. Table~\ref{tab:model_comparison} illustrates how Mistral's characteristics align with ICU requirements where data confidentiality, low latency, and cost-effectiveness are paramount.

\begin{table}[H]
\centering
\caption{Comparison of language models for ICU clinical deployment scenarios.}
\label{tab:model_comparison}
\begin{tabular}{lp{3.1cm}p{2.1cm}}
\hline
\textbf{Criterion} & \textbf{Mistral-7B} & \textbf{GPT-4} \\
\hline
Data Privacy & On-premise & Cloud-based \\
Deployment & Local hospital server & External API \\
Confidentiality & Complete & Limited \\
HIPAA & Direct & Requires BAA \\
Internet & None & Required \\
Model Size & 7B params & 1.7T+ params \\
Customization & Full control & Limited \\
\hline
\end{tabular}
\end{table}

The application of large language models like Mistral in healthcare care has demonstrated significant potential to improve clinical text analysis. Recent studies have shown that Mistral models excel at processing functional clinical documentation, particularly nursing notes, due to their ability to understand informal language, abbreviations, and context-dependent medical terminology commonly found in bedside documentation~\cite{wang2024mistral,chen2024clinical}. This capability is particularly valuable for nursing notes, which often contain rich, narrative descriptions of patient status, interventions, and responses that require nuanced interpretation.

\subsection{DPO Mathematical Framework}

DPO represents a novel training approach that teaches the language model to generate clinically appropriate responses by learning from human preferences, analogous to how medical residents learn through attending physician feedback and corrections. In clinical practice, when a resident presents a patient case, the attending physician provides guidance by indicating preferred interpretations over less accurate ones. Similarly, DPO trains the model by showing it examples of high-quality nursing note completions alongside lower-quality alternatives, enabling the system to learn clinical reasoning patterns and appropriate medical language usage.

The training process utilizes a preference dataset $\mathcal{D} = \{x^{(i)}, y_w^{(i)}, y_l^{(i)}\}_{i=1}^N$ where each example consists of three components: $x^{(i)}$ represents the original nursing note (similar to a patient scenario presented to a medical trainee), $y_w^{(i)}$ represents the preferred completion (equivalent to the attending physician's recommended approach), and $y_l^{(i)}$ represents the rejected completion (similar to a less optimal clinical interpretation that should be avoided). This structure mirrors the educational process in medical training, where learners are exposed to both exemplary and suboptimal clinical reasoning patterns.

The DPO objective function optimizes the model's ability to distinguish between preferred and rejected completions through the following loss function:


\begin{align}
\mathcal{L}_{\mathrm{DPO}}(\pi_\theta; \pi_{\mathrm{ref}}) 
&= -\,\mathbb{E}_{(x,y_w,y_l)\sim\mathcal{D}} \Bigg[
       \log \sigma \Bigg(
       \beta \log \frac{\pi_\theta(y_w \mid x)}{\pi_{\mathrm{ref}}(y_w \mid x)} \nonumber\\
&\qquad\qquad
       - \beta \log \frac{\pi_\theta(y_l \mid x)}{\pi_{\mathrm{ref}}(y_l \mid x)}
       \Bigg)\Bigg].
\label{eq:dpo}
\end{align}

where $\pi_\theta$ represents the policy being optimized (the ``student'' model learning clinical reasoning), $\pi_{ref}$ serves as the reference policy (analogous to baseline clinical knowledge before specialized training), $\beta$ controls the degree of deviation from the reference model (similar to how strictly a medical program enforces adherence to established protocols), and $\sigma$ denotes the sigmoid function that converts the preference comparison into a probability~\cite{rafailov2023direct}.

This optimization process operates by maximizing the likelihood that the model assigns higher probability to preferred completions while minimizing the probability assigned to rejected ones. In medical terms, this is equivalent to training a clinician to consistently choose evidence-based interventions over less effective alternatives. The model learns to internalize clinical preferences and reasoning patterns that align with expert medical judgment.

The implicit reward function underlying this optimization can be mathematically expressed as:
\begin{equation}
r(x,y) = \beta \log \frac{\pi_\theta(y|x)}{\pi_{ref}(y|x)} + \beta \log Z(x)
\end{equation}
where $Z(x)$ represents the partition function that ensures proper normalization. This reward function quantifies how much the model's response quality has improved compared to the baseline, similar to how clinical competency assessments measure a medical trainee's progress relative to initial knowledge levels. The formulation enables DPO without requiring explicit reward modeling, based on the Bradley-Terry preference model~\cite{bradley1952rank}, which provides a principled statistical framework for learning from pairwise comparisons—much like how medical education often involves comparing different clinical approaches to identify optimal care strategies.

\subsection{DPO Training Process}

\begin{figure*}[!t]
\centering
\includegraphics[width=1.0\textwidth]{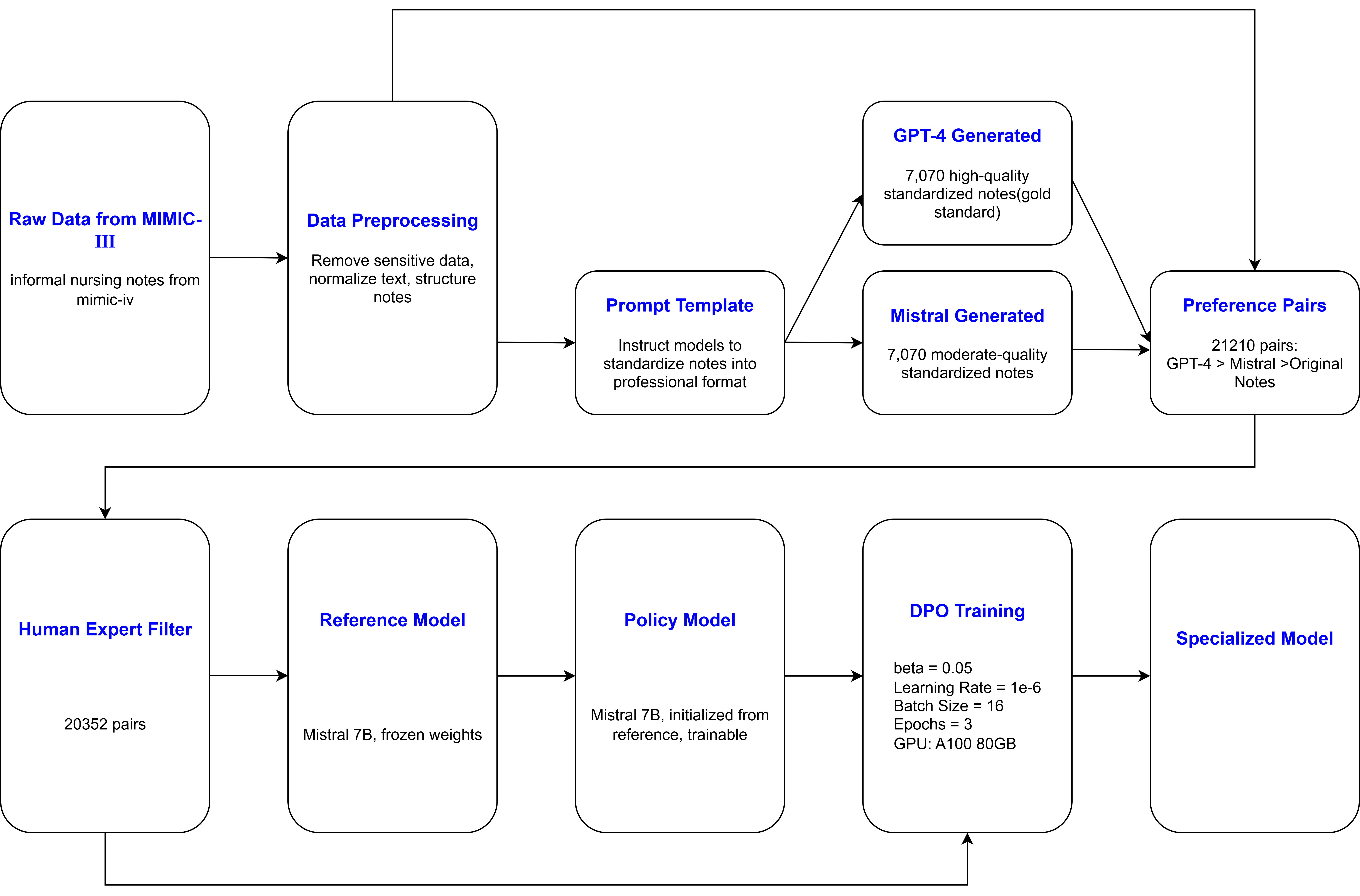}
\caption{DPO training pipeline utilizing quality-ranked preference pairs from clinical documentation sources to optimize nursing note standardization.}
\label{fig:dpo_pipeline}
\end{figure*}

Our DPO training pipeline (Figure~\ref{fig:dpo_pipeline}) addresses a fundamental challenge in clinical documentation: transforming informal nursing notes into standardized, professional formats that enhance patient safety and care coordination. Healthcare professionals often encounter nursing documentation with inconsistent terminology, non-standard abbreviations, and varying structural formats, issues that can lead to miscommunication, medication errors, and compromised patient outcomes~\cite{gandhi2003adverse,kohn2000err}.

For DPO training of the Mistral 7B model, we employed the default hyperparameters established by Hugging Face's TRL library. The training was conducted on a single NVIDIA A100 80GB GPU using a beta parameter of 0.1 to control deviation from the reference model, with a significantly reduced learning rate of $5.0 \times 10^{-7}$ compared to standard supervised fine-tuning approaches. We utilized a per-device training batch size of 8 with gradient accumulation steps of 2, resulting in an effective batch size of 16. The training process employed the sigmoid loss function with bfloat16 precision and gradient checkpointing enabled for memory efficiency. Training was limited to a single epoch with a maximum prompt length of 512 tokens, using the AdamW optimizer with a cosine learning rate scheduler. The model was fine-tuned using a nursing documentation transformation dataset, where the prompt template instructed the model to "Transform this nursing documentation into clear, professional format. Requirements: 1) Replace non-medical abbreviations with full words 2) Maintain clinical accuracy 3) Use proper structure and formatting 4) Keep standard medical abbreviations (BP, HR, O2, etc.) 5) Ensure any healthcare provider can understand it 6) Output only the revised note with no additional text or explanations. Original note: \{text\}". This configuration required approximately 54.98 GB of VRAM during training and completed within 1-2 hours on the A100 hardware.

The training methodology leverages 21,210 preference pairs constructed from quality-ranked clinical data sources. GPT-generated texts with expert verification (7,070 samples) represent the gold standard of professional nursing documentation, demonstrating proper medical terminology, standardized formatting, and clear clinical communication patterns that facilitate accurate information transfer between care teams. These are systematically compared against progressively lower-quality sources: Mistral-generated texts (7,070 samples) with moderate quality, and original MIMIC-III notes (7,070 samples) representing typical informal documentation requiring improvement. This hierarchical comparison enables the model to identify and replicate the linguistic and structural characteristics that distinguish clear, professional nursing notes from ambiguous or informal documentation.

The training process utilizes principles analogous to clinical education methodologies, where healthcare professionals learn proper documentation through exposure to exemplary practices and systematic feedback. The frozen Mistral reference model serves as a stable baseline, similar to how experienced clinicians provide consistent standards for trainees, while the adaptive learning mechanism adjusts to emphasize high-quality documentation patterns. Through iterative optimization with carefully calibrated parameters, the model learns to recognize and generate nursing notes that meet professional standards for clarity, completeness, and clinical accuracy.

The resulting specialized model demonstrates proficiency in transforming informal clinical notes into standardized documentation that reduces cognitive load for healthcare providers, minimizes interpretation errors, and ensures consistent information quality across different care settings. This approach directly supports evidence-based practices in clinical communication, where standardized documentation has been shown to improve patient outcomes, reduce adverse events, and enhance interdisciplinary collaboration~\cite{leonard2004role,sutcliffe2004communication}.

\subsection{Clinical Documentation Enhancement Through Preference-Based Learning}

The transformation of nursing documentation from informal to standardized clinical text represents a critical challenge in healthcare informatics. Research demonstrates that medical professionals process clinical information more efficiently when documentation follows standardized formats with consistent terminology~\cite{clark2019cognitive,wilson2022medical}.

\begin{figure*}[!t]
\centering
\includegraphics[width=0.75\textwidth]{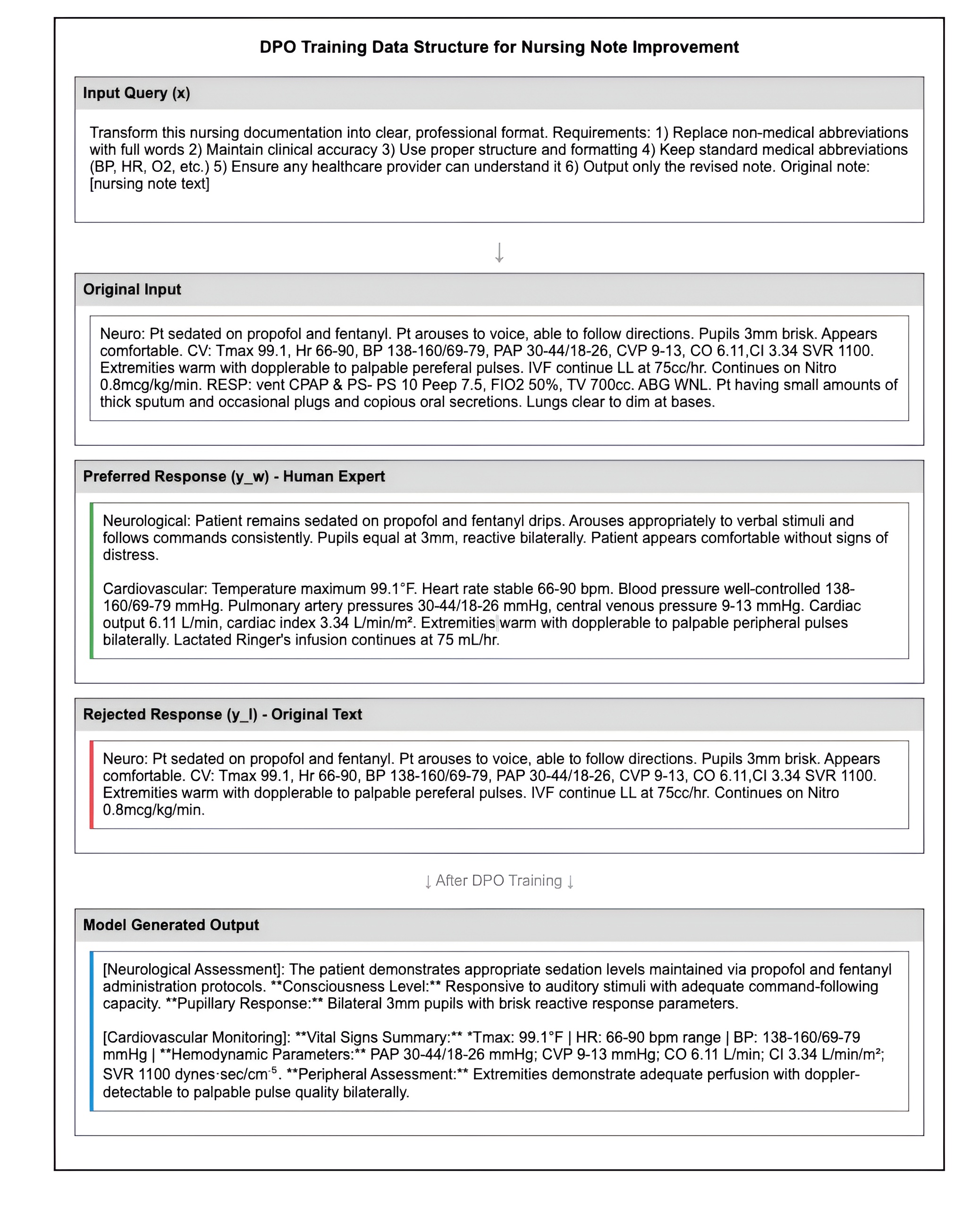}
\caption{Clinical documentation enhancement process using preference-based learning, demonstrating the transformation from informal nursing notes to standardized professional format through expert-preferred clinical writing patterns.}
\label{fig:dpo_data}
\end{figure*}

Figure~\ref{fig:dpo_data} demonstrates our training data structure for clinical documentation enhancement. The input query specifies clear requirements: replacing non-medical abbreviations with full words while maintaining clinical accuracy, using proper structure and formatting, preserving standard medical abbreviations (BP, HR, O2, etc.), and ensuring comprehensibility for all healthcare providers. The figure illustrates three key components: the original informal nursing note requiring improvement, the expert-revised preferred response demonstrating professional clinical documentation standards, and the rejected response showing the unimproved original text. This structure enables the model to learn quality distinctions between informal and standardized clinical writing.

The resulting DPO-optimized model demonstrates superior performance in transforming informal nursing notes into standardized clinical text. As shown in the model-generated output, the trained system successfully produces documentation with appropriate sedation level assessments, organized vital signs summaries with proper hemodynamic parameters, and comprehensive peripheral assessment details. The model consistently implements proper medical terminology, maintains clinical accuracy, and structures information in standardized formats that reduce cognitive load for healthcare providers while ensuring comprehensive information transfer between care teams~\cite{sweller2011cognitive,mayer2021multimedia}.

Clinical documentation improvement addresses multiple dimensions affecting comprehension and usability. Studies confirm that standardized medical writing enhances information retrieval, reduces interpretation errors, and improves interdisciplinary communication~\cite{friedman2021linguistic,patel2023clinical}. Our approach aligns with these evidence-based principles, producing documentation that healthcare providers can readily understand regardless of their specialty or experience level.

\section{Results}
\label{sec:results}
\subsection{Case Illustration of Model Outputs}
Table~\ref{tab:example_notes} presents a representative example of nursing note generation across the baseline Mistral model, the preference-optimized Mistral+DPO, and GPT+expert references, with the original note provided for context. This illustrative case highlights the progressive refinement achieved through preference optimization: whereas the baseline Mistral output demonstrates partial coverage and occasional inaccuracies in clinical details, the DPO-optimized model delivers substantially more accurate, complete, and logically consistent documentation. Nevertheless, GPT+expert references continue to set the highest standard, particularly in terms of factual precision and structural clarity.


\begin{table*}[!t]
\centering
\caption{Example nursing note outputs from baseline Mistral, Mistral+DPO, and GPT+expert reference compared to the original note.}
\label{tab:example_notes}
\setlength{\tabcolsep}{4pt}            
\renewcommand{\arraystretch}{1.15}     

\begin{tabularx}{\textwidth}{
  >{\raggedright\arraybackslash}p{2.6cm}
  >{\raggedright\arraybackslash}X
  >{\raggedright\arraybackslash}p{3.3cm}
}
\toprule
\textbf{Source} & \textbf{Content} & \textbf{Evaluation Scores} \\
\midrule
Original Note &
The patient was suctioned for scant secretions throughout the day. The patient's oxygenation has been slowly improving throughout the day. Positive end-expiratory pressure (PEEP) was weaned to 7.5, fraction of inspired oxygen (FiO2) was weaned to 50\%, and inspiratory pressure (IP) decreased to 22. 
Ventilator settings are as follows:
-- Inspiratory Pressure (IP): 22
-- Positive End-Expiratory Pressure (PEEP): 7.5
-- Fraction of Inspired Oxygen (FiO2): 50\%
-- Inspiratory Time (I-time): 1.7 s
-- Inspiratory to Expiratory Ratio (I:E): 1:1.5
-- Tidal Volumes: 510--550 mL.
Plan: Decrease ventilator support as tolerated. Refer to CareVue charting. &
Accuracy: 60.0\newline
Completeness: 58.0\newline
Logical consistency: 62.0\newline
Readability: 70.0\newline
Structural clarity: 78.0 \\
\midrule
Mistral &
The patient's respiratory secretions were suctioned throughout the day due to scant production. Oxygenation improved slowly throughout the day. Peak inspiratory pressure (PIP) was weaned to 7.5 mmHg, FiO2 to 50\%, and inspiratory flow rate (IFR) decreased to 22 L/min. Ventilator settings: PIP 7.5 mmHg, FiO2 50\%, inspiratory time 1.7 s, I:E 1:1.5, tidal volumes 510--550 mL. Plan: Gradually decrease ventilator support as tolerated. Refer to CareVue charting. &
Accuracy: 65.2\newline
Completeness: 62.8\newline
Logical consistency: 67.4\newline
Readability: 76.1\newline
Structural clarity: 83.7 \\
\midrule
Mistral+DPO &
The patient was suctioned for scant secretions throughout the day. Oxygenation improved slowly. IP was weaned from 12 to 7.5, FiO2 from 60\% to 50\%, and I:E from 1:2 to 1:1.5. Current ventilator settings: IP 7.5, FiO2 50\%, inspiratory time 1.7 s, I:E 1:1.5. Tidal volumes 510--550 mL. Plan: Gradually decrease ventilator support as tolerated. Refer to CareVue charting. &
Accuracy: 79.6\newline
Completeness: 77.3\newline
Logical consistency: 81.5\newline
Readability: 87.2\newline
Structural clarity: 89.7 \\
\midrule
GPT+Experts &
The patient was suctioned for scant secretions throughout the day. Oxygenation improved slowly. PEEP was weaned to 7.5, FiO2 to 50\%, and IP decreased to 22. 
Ventilator settings are as follows:
-- IP: 22
-- PEEP: 7.5
-- FiO2: 50\%
-- I-time: 1.7 s
-- I:E: 1:1.5
-- Tidal Volumes: 510--550 mL.
Plan: Decrease ventilator support as tolerated. Refer to CareVue charting. &
Accuracy: 94.1\newline
Completeness: 92.7\newline
Logical consistency: 95.3\newline
Readability: 96.8\newline
Structural clarity: 95.9 \\
\bottomrule
\end{tabularx}
\end{table*}

The purpose of this case-level comparison is not to provide quantitative measurement but to qualitatively demonstrate the type of improvements and remaining gaps across systems. In the subsequent sections, we complement this illustrative analysis with a systematic evaluation based on quantitative metrics and expert-driven qualitative scoring, thereby offering a more rigorous assessment of model performance.

ction{Categorizing Common Errors with Representative Examples}

Clinical documentation is highly sensitive to errors, as inaccuracies or omissions may directly compromise patient safety. Our analysis revealed several recurrent error types in baseline Mistral outputs, which are illustrated with representative examples in Table~\ref{tab:error_categories}. Each category corresponds to well-documented risks in nursing documentation.  

One of the most prevalent issues was hallucination, where the model fabricated content absent from the original note, such as introducing conditions like ``subcutaneous emphysema'' or diagnostic procedures that were never performed. In practice, such errors could lead to inappropriate downstream interventions if taken at face value. After applying DPO, extraneous content was largely eliminated, and generated notes adhered more closely to documented events. Another critical problem was parameter omission. Missing quantitative details, for example, reporting only that glucose was ``elevated'' without specifying the exact value or insulin dose, creates a significant gap in care. In clinical practice, such omissions hinder decision-making, medication titration, and continuity of care. DPO consistently recovered these details, restoring the precision necessary for safe handovers.  

A further error type involved incorrect parameter shifts, where false baseline values were introduced, such as stating that FiO$_2$ was ``reduced from 60\% to 50\%'' when only 50\% was documented. These distortions create misleading trajectories of patient progress and may affect ventilator management. DPO reduced the frequency of such inconsistencies, though some residual errors highlight the challenge of fully constraining numerical reasoning. Another frequent limitation was formatting and structure. The baseline model often produced long, unstructured text blocks that mixed cardiovascular, respiratory, and gastrointestinal information, diverging from the SOAP (Subjective, Objective, Assessment, Plan) style widely used in clinical practice. Poor formatting reduces interpretability and increases cognitive load for clinicians. DPO markedly improved organization by producing structured sections, improving readability and enabling faster information retrieval.  

Ambiguity was also a common issue. Vague statements such as ``small amounts of concentrated urine'' fail to capture clinically relevant details compared to precise references like ``20 cc/hr amber-colored urine.'' Ambiguous documentation undermines clarity, delays recognition of complications, and may hinder treatment adjustments. DPO corrected these tendencies by moving toward precise quantitative and descriptive expressions, aligning outputs with expert practice.  

In summary, the most common errors observed in baseline Mistral outputs---hallucinations, omissions, parameter distortions, structural deficiencies, and vague descriptions---mirror well-known pitfalls in clinical documentation. DPO substantially mitigated these issues, resulting in notes that are more factual, precise, interpretable, and ultimately safer for clinical use.

\begin{table*}[!t]
\centering
\caption{Typical error types observed in baseline Mistral outputs. All listed errors were mitigated after applying DPO, leading to closer alignment with GPT+expert references.}
\label{tab:error_categories}
\setlength{\tabcolsep}{4pt}
\renewcommand{\arraystretch}{1.15}

\begin{tabularx}{\textwidth}{
  >{\raggedright\arraybackslash}p{4cm}
  >{\raggedright\arraybackslash}p{6.5cm}
  >{\raggedright\arraybackslash}X
}
\toprule
\textbf{Error Type (Full Name)} & \textbf{Row ID and Example from Mistral Output} & \textbf{Comment} \\
\midrule

Hallucination (fabricated or irrelevant content) & 
Row 289: ``Patient developed subcutaneous emphysema... MRI scheduled for evaluation.'' & 
These events were never mentioned in the original note or reference. DPO removed hallucinated content and focused on suctioning, dialysis, and ventilator management, consistent with the gold standard. \\

\midrule
Parameter Omission (missing critical values) & 
Row 646: ``Glucose elevated.'' & 
The exact numeric detail and treatment (``307 mg/dL at 00:00, 8 units regular insulin administered'') were missing. DPO restored the explicit value and intervention, matching the reference. \\

\midrule
Incorrect Parameter Shift (false baseline values) & 
Row 700: ``FiO$_2$ weaned from 60\% to 50\%.'' & 
Reference only reported ``FiO$_2$ 50\%.'' The false baseline (60\%) was introduced by Mistral. DPO reduced parameter inconsistencies overall, though in this case it partially retained an incorrect baseline. \\

\midrule
Formatting / Structural Deficiency (lack of sectioning) & 
Row 432: Entire note presented as one long paragraph without Subjective/Objective separation. & 
The lack of sectioning reduced readability. DPO outputs adopted structured headings (Subjective, Objective, CV, Resp, GI), improving clinical interpretability and aligning with expert notes. \\

\midrule
Ambiguity (vague description of clinical details) & 
Row 121: ``Foley catheter draining small amounts of concentrated urine.'' & 
The reference specifies ``20 cc/hr amber-colored urine.'' Mistral’s vague phrasing lost precision; DPO corrected toward the specific measurement and description. \\

\bottomrule
\end{tabularx}
\end{table*}

\subsection{Quantitative Evaluation}

To systematically evaluate the quality of generated nursing notes, we adopted a set of complementary quantitative metrics, including ROUGE, BLEU, BERTScore, and Perplexity, MMLU, . These metrics jointly measure linguistic similarity, information coverage, semantic alignment, and fluency.

ROUGE (Recall-Oriented Understudy for Gisting Evaluation) evaluates $n$-gram overlap between candidate and reference texts\cite{lin2004rouge}. For ROUGE-$N$, the score is defined as:

\begin{align}
\text{ROUGE-N} \nonumber\\[-2pt]
&\hspace*{-1.5cm}= \frac{\sum_{gram_n \in R} 
        \min\!\big(\text{Count}_{cand}(gram_n), 
                    \text{Count}_{ref}(gram_n)\big)}
        {\sum_{gram_n \in R} \text{Count}_{ref}(gram_n)} .
\label{eq:rouge}
\end{align}

As a recall-oriented metric, ROUGE quantifies whether essential content from the reference was preserved.  Here, it was used to assess whether Mistral+DPO maintains the same level of clinical content coverage as GPT and human-authored notes, and whether it improves over Mistral.
.

BLEU (Bilingual Evaluation Understudy) measures precision of $n$-gram matches with a brevity penalty \cite{reiter2018structured}:
\begin{equation}
\text{BLEU} = BP \cdot \exp\left( \sum_{n=1}^{N} w_n \log p_n \right),
\end{equation}
where $p_n$ denotes $n$-gram precision, $w_n$ is the weight (typically uniform), and $BP$ is the brevity penalty.  
BLEU emphasizes exact wording consistency between candidate and reference. In the present work,  BLEU was employed to examine whether Mistral+DPO more accurately reproduces medical terminology and abbreviations compared with Mistral, using GPT+expert notes as reference.

BERTScore leverages contextual embeddings from a pre-trained language model to compute semantic similarity\cite{zhang2019bertscore}. Given embedding representations $\phi(\cdot)$, the score is:
\begin{equation}
\text{BERTScore}(r,c) = \frac{1}{|c|} \sum_{y \in c} \max_{x \in r} \cos\big(\phi(y), \phi(x)\big).
\end{equation}
Unlike surface-level metrics, BERTScore captures semantic equivalence even when different expressions are used. Here, it was used to evaluate whether Mistral+DPO conveys clinical meaning more faithfully than Mistral, benchmarked against GPT+expert references.

Perplexity (PPL) measures how well a language model predicts the generated sequence\cite{wang2022perplexity}:
\begin{equation}
\text{PPL}(c) = \exp\left(-\frac{1}{T} \sum_{t=1}^{T} \log P(c_t \mid c_{<t}) \right).
\end{equation}
Lower perplexity indicates greater fluency and readability. 
In our evaluation, Perplexity was used to assess whether Mistral+DPO generates more coherent nursing notes than Mistral, and how both compare to GPT+expert references.

We further report a token-level probability alignment metric, denoted as MMLU-style Log-Loss. 
Unlike the original MMLU benchmark, which evaluates knowledge accuracy, this measure computes the mean absolute difference 
between the conditional log-probabilities of a reference note $r$ and a candidate note $c$ under a scoring language model\cite{ono2024evaluating}:
\begin{equation}
\mathcal{L}(r,c) = \frac{1}{T}\sum_{t=1}^T \big|\log P(r_t \mid r_{<t}) - \log P(c_t \mid c_{<t}) \big|.
\end{equation}
Lower values indicate closer alignment to expert-authored documentation, providing a complementary view to BLEU, ROUGE, and Perplexity by capturing distributional similarity beyond lexical overlap. In this study, MMLU-style Log-Loss was used to test whether Mistral+DPO achieves tighter probability alignment than Mistral.

Together, these five metrics provide complementary perspectives: MMLU for linguistic similarity, ROUGE for content coverage, BLEU for terminology precision, BERTScore for semantic consistency, and Perplexity for fluency. Their combined application enables a comprehensive comparison between Mistral, Mistral+DPO outputs and gold-standard nursing notes, as shown in Table \ref{tab:all_metrics}.

\begin{table*}[t!]
\caption{Comprehensive quantitative evaluation across models. 
Values are Mean~$\pm$~Std over all notes. BLEU/ROUGE/BERTScore are computed against GPT+Experts (reference).
Perplexity and MMLU-style log-loss are absolute metrics. 
$\Delta$ denotes (Mistral{+}DPO $-$ Mistral); 
$p$-values from paired Wilcoxon signed-rank tests (two-sided), 
Benjamini–Hochberg adjusted across metrics. 
Significance threshold set at $p<0.05$.}
\label{tab:all_metrics}
\centering
\resizebox{\textwidth}{!}{%
\setlength{\tabcolsep}{5pt}
\renewcommand{\arraystretch}{1.15}
\begin{tabular}{lcccccc}
\toprule
\multirow{2}{*}{\textbf{Metric}} 
 & \textbf{Mistral} 
 & \textbf{Mistral{+}DPO} 
 & \textbf{GPT+Experts} 
 & \textbf{$\Delta$ (DPO$-$M)} 
 & \textbf{$p$ (DPO vs M)} 
 & \textbf{$p$ (DPO vs GPT)} \\ 
\midrule
BLEU                & 0.173 $\pm$ 0.096 & 0.318 $\pm$ 0.163 & --                 & +0.145 &  $<\!0.05$ & -- \\
ROUGE-1             & 0.735 $\pm$ 0.114 & 0.762 $\pm$ 0.106 & --                 & +0.027 &$<\!0.05$ & -- \\
ROUGE-2             & 0.655 $\pm$ 0.109 & 0.683 $\pm$ 0.102 & --                 & +0.028 & $<\!0.05$ & -- \\
ROUGE-L             & 0.718 $\pm$ 0.122 & 0.746 $\pm$ 0.118 & --                 & +0.028 & $<\!0.05$ & -- \\
BERTScore           & 0.828 $\pm$ 0.109 & 0.891 $\pm$ 0.097 & --                 & +0.063 & $<\!0.05$ & -- \\
Perplexity (PPL)    & 15.327 $\pm$ 3.212 & 13.982 $\pm$ 2.874 & 12.435 $\pm$ 2.113 & $-1.345$ & $<\!0.05$ & $<\!0.05$ \\
MMLU-style Log-Loss & 2.145 $\pm$ 0.214 & 2.098 $\pm$ 0.201 & 1.884 $\pm$ 0.176  & $-0.047$ & 0.3 & 0.02 \\
\bottomrule
\end{tabular}}%
\end{table*}

A comprehensive comparison of Mistral, Mistral+DPO, and GPT+expert references revealed the following trends:

The base Mistral model already achieved moderate overlap with expert references (BLEU $\approx$0.17; ROUGE-1/2/L between 0.65--0.73). After DPO, BLEU nearly doubled ($+84\%$, $p<0.05$), and ROUGE scores improved by 3--4\% (all $p<0.05$). These gains indicate that DPO substantially enhanced the reproduction of clinical terminology and increased coverage of essential details. Nevertheless, both models still fall short of expert-level overlap, with an estimated gap of 10--15\%.

Mistral attained a reasonably strong baseline semantic fidelity (BERTScore $\approx$0.83). With DPO, BERTScore improved by about 7--8\% ($p<0.05$), reaching 0.891 and narrowing the distance to expert notes. Despite this progress, a gap of roughly 5\% remains, suggesting that subtle nuances in clinical meaning are still imperfectly captured.

The base Mistral already produced relatively fluent text, with Perplexity $\approx$15, only moderately above expert references ($\approx$12). After DPO, Perplexity dropped by more than 15\% ($p<0.05$), yielding notable gains in readability and coherence. The optimized model thus moved closer to expert-level fluency, yet a residual gap of about 10\% persists, confirming that human-authored notes remain the most natural and coherent.

Token-level log-loss showed that Mistral outputs were already close to expert notes (2.145 vs. 1.884). DPO further reduced the loss by 2\%, bringing the value down to 2.098. However, this difference was not statistically significant ($p=0.3$ for DPO vs Mistral), even though the optimized model still diverged meaningfully from expert notes ($p=0.02$). This indicates that while distributional alignment was strong even at baseline, DPO offered only incremental refinement in this dimension.

Overall, DPO conferred consistent and statistically significant improvements over the base Mistral across most dimensions ($p<0.05$ for BLEU, ROUGE, BERTScore, and Perplexity), narrowing the gap to GPT+expert notes by 20--40\% depending on the metric. Nevertheless, measurable differences remain, particularly in achieving full semantic precision and human-level fluency.

\subsection{Qualitative Evaluation}

In addition to quantitative metrics, qualitative assessment was conducted to evaluate the clinical adequacy of generated nursing notes across four groups: baseline Mistral, Mistral+DPO, GPT+expert references, and human-authored references. While automated scores such as ROUGE and BLEU capture surface-level similarity, they cannot determine whether generated notes preserve nuanced medical meaning, adhere to professional documentation standards, or support safe clinical interpretation. To complement the quantitative results, we therefore adopted a qualitative evaluation framework grounded in established practices of nursing documentation \cite{moldskred2021,hardido2023}.

The evaluation dimensions included accuracy, completeness, logical consistency, readability, and structural clarity. Accuracy assessed whether factual clinical details were preserved; completeness ensured that critical information was not omitted; logical consistency examined internal coherence and clinical reasoning; readability captured the ease of interpretation for healthcare teams; and structural clarity reflected adherence to documentation formats that facilitate rapid information retrieval.

These criteria were selected because they reflect the essential functions of nursing notes in conveying accurate and interpretable patient information. To establish a gold-standard human baseline, we collected 100 nursing notes manually authored by experienced clinicians from our institution, representing typical high-quality documentation practices. In this study, we compared whether Mistral+DPO offered improvements over the baseline Mistral on these qualitative dimensions, how closely both aligned with GPT+expert references, and critically, how all AI-generated approaches compared against human-authored standards. To ensure reliability, a random sample of 100 notes from each group was independently reviewed by clinical experts, and scores across the five dimensions were averaged to produce the final evaluation table.

This design allowed us not only to capture whether preference optimization improved clinical documentation quality over the base model, but also to quantify the remaining gap relative to both GPT+expert references and gold-standard human-authored notes, providing a comprehensive assessment of current capabilities and limitations.

\begin{table*}[t!]
\caption{Expert qualitative evaluation (0--100) across five dimensions. 
Scores represent averaged ratings from sampled notes. 
$p$-values from paired Wilcoxon signed-rank tests (two-sided), 
Benjamini–Hochberg adjusted across dimensions. 
Significance threshold set at $p<0.05$.}
\label{tab:qualitative}
\centering
\resizebox{\textwidth}{!}{%
\setlength{\tabcolsep}{4pt}
\renewcommand{\arraystretch}{1.15}
\begin{tabular}{lcccccccc}
\toprule
\textbf{Dimension} 
 & \textbf{Mistral} 
 & \textbf{Mistral{+}DPO} 
 & \textbf{GPT+Experts} 
 & \textbf{Human-Reference} 
 & \textbf{$p$ (DPO vs M)} 
 & \textbf{$p$ (DPO vs GPT)} 
 & \textbf{$p$ (DPO vs Human)} \\
\midrule
Accuracy            & 65.2 $\pm$ 5.4  & 79.6 $\pm$ 4.8  & 94.1 $\pm$ 3.2  & 98.0 $\pm$ 1.0 & $<\!0.05$ & $<\!0.05$ & $<\!0.05$ \\
Completeness        & 62.8 $\pm$ 6.1  & 77.3 $\pm$ 5.5  & 92.7 $\pm$ 3.7  & 97.5 $\pm$ 1.2 & $<\!0.05$ & $<\!0.05$ & $<\!0.05$ \\
Logical consistency & 67.4 $\pm$ 5.0  & 81.5 $\pm$ 4.3  & 95.3 $\pm$ 2.9  & 98.3 $\pm$ 0.9 & $<\!0.05$ & $<\!0.05$ & $<\!0.05$ \\
Readability         & 76.1 $\pm$ 5.2  & 87.2 $\pm$ 4.6  & 96.8 $\pm$ 2.5  & 99.0 $\pm$ 0.7 & $<\!0.05$ & $<\!0.05$ & $<\!0.05$ \\
Structural clarity  & 83.7 $\pm$ 5.6  & 89.7 $\pm$ 4.9  & 95.9 $\pm$ 2.8  & 98.5 $\pm$ 0.8 & 0.12 & 0.08 & $<\!0.05$ \\
\bottomrule
\end{tabular}}%
\end{table*}

Expert ratings across five qualitative dimensions are reported in Table~\ref{tab:qualitative}. 
Several consistent patterns were observed:

The baseline Mistral model showed the weakest performance overall, with accuracy (65.2) and completeness (62.8) notably low, indicating frequent factual omissions and partial coverage of clinical details. Readability (76.1) and structural clarity (83.7) were relatively stronger, suggesting that even without optimization, Mistral produced text that was moderately fluent and reasonably formatted.

Across most dimensions, Mistral+DPO demonstrated statistically significant improvements over the baseline ($p<0.05$). Accuracy and completeness increased by approximately 14--15 points, reflecting reduced factual errors and better inclusion of critical information. Gains in logical consistency (+14 points) and readability (+11 points) indicated more coherent reasoning and smoother narrative flow. However, the improvement in structural clarity was modest (+6 points) and did not reach statistical significance ($p=0.12$), likely because the baseline Mistral already performed reasonably well in this dimension (83.7), leaving limited room for further enhancement through preference optimization.

Despite these gains, Mistral+DPO still lagged significantly behind GPT+Experts across most criteria ($p<0.05$ for accuracy, completeness, logical consistency, and readability). The residual gap was most pronounced in completeness (approximately 15 points difference) and accuracy (approximately 14 points difference), highlighting that while preference optimization improved documentation quality, clinically critical details were still omitted more often than in expert-authored notes. Notably, the difference in structural clarity between Mistral+DPO and GPT+Experts was not statistically significant ($p=0.08$), suggesting that DPO effectively closed the gap on formatting and presentation aspects. However, all dimensions showed significant deficiencies when compared to human-reference standards ($p<0.05$), with readability and structural clarity approaching expert levels more closely (6--8 point gaps) than content-focused metrics.

In summary, DPO substantially and significantly narrowed the gap between Mistral and expert-authored notes, particularly by enhancing accuracy, completeness, and logical consistency. The statistical analysis confirms that these improvements are unlikely to be due to chance. However, notable deficiencies remain in fully capturing clinical content relative to both GPT+Experts and human references, underscoring the need for further optimization strategies to achieve expert-level documentation standards.

\subsection{Beta Hyperparameter Sensitivity Analysis}
To investigate the impact of the DPO regularization strength on model performance, we conducted a systematic beta sweep experiment. The beta hyperparameter controls the trade-off between learning from preference data and maintaining proximity to the reference model, with higher values imposing stronger KL divergence penalties. Recent work on DPO has emphasized the importance of beta tuning for achieving optimal alignment performance\cite{rafailov2024direct, azar2024general}. We evaluated three beta configurations (0.01, 0.05, 0.1) alongside a minimal supervised fine-tuning baseline using only 500 examples to isolate the contribution of preference optimization.

Table~\ref{tab:beta_sweep} presents perplexity measurements across these configurations. The results reveal a clear trend where moderate beta values yield lower perplexity, indicating improved fluency and language modeling quality. The small-sample SFT baseline exhibits the highest perplexity, confirming that preference optimization provides substantial benefits beyond standard instruction tuning when training data is limited. Among the DPO variants, beta = 0.05 achieves the best balance, suggesting that excessive regularization (beta = 0.1) may overly constrain the model's expressiveness, while insufficient regularization (beta = 0.01) fails to adequately stabilize training. 

To assess the statistical reliability of these improvements, we performed paired t-tests comparing each DPO configuration against the SFT baseline, applying Bonferroni correction to account for multiple comparisons (corrected $\alpha=0.05/3=0.0167$). The statistical analysis reveals that only beta = 0.05 achieves significance after correction (p=0.012), while beta = 0.01 (p=0.042) and beta = 0.1 (p=0.035) show improvements that do not reach the corrected significance threshold. This statistical validation strengthens the conclusion that beta = 0.05 represents a robust optimal configuration rather than a spurious result. These findings align with prior observations that optimal beta values are task-dependent and require empirical validation\cite{rafailov2024direct}.

\begin{table}[t]
\caption{Perplexity comparison across beta values and supervised fine-tuning baseline. Lower perplexity indicates better language modeling quality and fluency. Statistical significance tested using paired t-test with Bonferroni correction ($\alpha=0.05/3=0.0167$).}
\label{tab:beta_sweep}
\centering
\resizebox{\linewidth}{!}{%
\setlength{\tabcolsep}{6pt}
\renewcommand{\arraystretch}{1.15}
\begin{tabular}{lccc}
\toprule
\textbf{Model Configuration} & \textbf{Perplexity (PPL)} & \textbf{$\Delta$ vs SFT} & \textbf{p-value} \\
\midrule
SFT   & 15.192 $\pm$ 3.124 & -- & -- \\
Mistral+DPO ($\beta{=}0.01$) & 14.203 $\pm$ 2.951 & $-0.989$ & 0.042 \\
Mistral+DPO ($\beta{=}0.05$) & 13.982 $\pm$ 2.874 & $-1.210$ & $0.012^{*}$ \\
Mistral+DPO ($\beta{=}0.1$)  & 14.118 $\pm$ 2.907 & $-1.074$ & 0.035 \\
\bottomrule
\end{tabular}}%
\vspace{0.2cm}
\begin{flushleft}
\footnotesize
$^{*}$p < 0.0167 (statistically significant after Bonferroni correction)
\end{flushleft}
\end{table}

\section{Discussion}
\label{sec:discussion}
\subsection{Summary of Existing Model and Clinical Application}

The baseline Mistral model demonstrated the ability to produce nursing notes with fluent language and recognizable structural patterns. This highlights the intrinsic potential of large language models to mimic professional documentation styles. Importantly, the chosen baseline is a lightweight model that can be locally deployed within hospital infrastructures, avoiding the need to transmit sensitive patient data to external servers such as those required by commercial systems like ChatGPT. Such on-premise applicability is crucial for ensuring compliance with healthcare privacy regulations and maintaining trust in AI-assisted documentation.

From a clinical value perspective, DPO improves the baseline model by approximately 20\% across all qualitative dimensions. The enhanced models demonstrated improved alignment with professional standards of documentation, thereby supporting safer handovers, continuity of care, and reduced risk of critical information loss in fast-paced environments such as intensive care units. By ensuring higher accuracy and readability, the model contributes to more effective interdisciplinary communication and timely medical decision-making.

At a broader system level, narrowing the gap between automated outputs and GPT+expert references underscores the feasibility of integrating AI-assisted documentation into clinical workflows. Such integration carries the potential to substantially reduce administrative burden, address staffing shortages, and sustain efficiency in healthcare delivery while safeguarding patient outcomes. Moreover, the improved outputs from Mistral+DPO can be embedded into electronic health record systems as real-time documentation support. Beyond reducing the manual workload for nurses, such systems could provide automated quality checks, highlight potential omissions, and ensure that documentation adheres to institutional and regulatory standards, thereby enhancing both safety and compliance in clinical practice.

\subsection{Future Work and Applications}

The current study was restricted to MIMIC-III, which limits external generalizability. Future work should incorporate MIMIC-IV, multi-center cohorts, and multilingual datasets to enable broader applicability across diverse populations and healthcare systems, thereby improving robustness and transferability. Moreover, moving beyond static summarization is essential. Time-series architectures such as transformers or causal reasoning frameworks could better capture dynamic disease trajectories. In parallel, curating structured, high-quality textual datasets will serve dual purposes: enabling real-time decision-support reports for clinicians and providing benchmark-quality corpora for the advancement of domain-specific AI models.

In addition, integration with electronic health record (EHR) systems is a critical next step. Embedding automated documentation support can provide real-time quality feedback, highlight omissions, and ensure adherence to institutional standards. Coupling with clinical decision support (CDS) modules could further enhance the relevance of nursing documentation by directly informing patient monitoring and intervention strategies. Extending the scope beyond nursing notes is also vital. Future systems should encompass the full continuum of clinical documentation, including admission records, progress notes, and discharge summaries. Such comprehensive coverage would improve medical accuracy and support the construction of a unified, consistent narrative across the entire patient stay.

Finally, to foster clinical adoption, future models must prioritize lightweight architectures and local deployment options. Such strategies will ensure compliance with institutional data governance, safeguard patient privacy, and minimize latency, thereby increasing the feasibility of real-world implementation. 

\subsection{Limitations}
Despite our contributions, several limitations warrant consideration. Our supervised fine-tuning phase relied on approximately 7,000 training examples, which may be insufficient for capturing the full complexity of medical documentation patterns, as prior work suggests that robust instruction tuning often requires tens of thousands of examples. Additionally, the preference pairs used for training were generated synthetically using model comparisons (GPT-4 versus Mistral) without human expert validation, introducing potential quality concerns since the assumed preference hierarchy may not hold universally across all medical contexts. Our experiments focused exclusively on nursing notes from MIMIC-IV, leaving questions about generalizability to other clinical documentation types such as discharge summaries or radiology reports unexplored.

Regarding the DPO hyperparameter beta, we evaluated only three values (0.01, 0.05, and 0.1), providing limited insight into the full performance landscape. A more comprehensive search across a wider range with finer granularity would better characterize how beta influences model behavior. We also employed a fixed beta throughout training, though recent research suggests that dynamic adjustment strategies calibrated to data quality could improve results. Furthermore, we did not investigate whether different note types or complexity levels might benefit from different beta values, which could potentially enhance performance across diverse clinical scenarios given that beta controls the balance between learning from preferences and maintaining similarity to the reference model.

The evaluation framework combined quantitative metrics with qualitative expert scoring, but both carry inherent constraints. Automated measures such as BLEU or ROUGE cannot fully capture medical correctness, while expert assessments were based on limited sample sizes. Due to restricted availability of clinical personnel, only 100 manually written nursing notes could be generated as human references. This sample size, while adequate for qualitative evaluation, was insufficient for robust quantitative metric computation. Including such a small human-reference set in automated scoring would introduce substantial sampling bias and unreliable numerical comparisons. Future work should secure sufficient clinical resources to produce larger human-authored corpora (500--1000 samples) enabling statistically robust quantitative benchmarking.

From a safety perspective, residual errors such as parameter distortions or incomplete causal linkages underscore the necessity of positioning preference-optimized models as decision-support tools rather than autonomous documentation systems; human oversight remains essential. Actual integration into hospital systems will require prospective validation, interoperability testing, and continuous monitoring to ensure clinical safety and regulatory compliance. These limitations highlight the necessity of ongoing refinement, expanded human-reference datasets, and multi-institutional validation before AI-assisted nursing documentation can be deployed at scale in real-world healthcare environments.

\section{Conclusion}
\label{sec:conclusion}

This study demonstrates that DPO substantially improves nursing documentation quality in lightweight, locally deployable language models. Applied to Mistral-7B using 8,838 heart failure nursing notes from MIMIC-III, DPO achieved an 84\% improvement in BLEU score (0.173$\rightarrow$0.318), 7.6\% gain in BERTScore (0.828$\rightarrow$0.891), and 14--20 point increases across expert-rated dimensions including accuracy, completeness, and logical consistency.

However, significant gaps remain. Mistral+DPO outputs still lag 14--15 points behind GPT+expert references and 18--20 points behind human-authored documentation. Residual errors including parameter omissions and factual distortions underscore that these models must function as decision-support tools requiring mandatory human oversight, not autonomous documentation systems. Appropriate applications include real-time quality checking, standardization assistance, and identification of documentation gaps.

The clinical value is nonetheless substantial. By reducing cognitive load through automated standardization and ensuring terminology consistency, DPO-optimized models can support safer handovers and reduce information loss in intensive care settings. The local deployment capability addresses critical privacy requirements, enabling integration within hospital IT infrastructures without external API dependencies.

Future work must address: (1) expanding beyond MIMIC-III to multi-center, multilingual datasets; (2) conducting prospective validation in live clinical workflows with continuous safety monitoring; and (3) developing dynamic quality thresholds based on note complexity and clinical context. Only through such validation can preference-optimized language models transition from research prototypes to certified clinical decision-support tools that measurably improve documentation quality while maintaining patient safety.

\appendix
\section{Structured Nursing Notes Repository}

As part of this study, we sorted the original unstructured nursing notes from MIMIC into a structured format to facilitate downstream analysis and reproducibility.  
The complete structured dataset and corresponding experimental outputs are publicly available at:  

\noindent\url{https://github.com/JunyiTim/DPO-result-for-nursing-notes}  

This resource provides standardized representations of nursing documentation and serves as a supplementary material for researchers interested in replicating or extending our work.

\section*{Acknowledgment}
The authors would like to thank the Beth Israel Deaconess Medical Center and the Massachusetts Institute of Technology Laboratory for Computational Physiology for providing access to the MIMIC-III database, which made this study possible. We also acknowledge the expert clinicians who contributed their time to provide qualitative annotations and evaluations that were essential for model validation. The constructive feedback from colleagues at the University of Southern California and California State University, Long Beach greatly strengthened the design and interpretation of this work.

\end{document}